\documentclass[11pt]{article}

\usepackage[T1]{fontenc}
\usepackage[margin=1in]{geometry}
\usepackage{amsmath,amssymb}
\usepackage{graphicx}
\usepackage{enumitem}
\usepackage{xcolor}
\usepackage{tikz}
\usepackage{adjustbox}
\usepackage{booktabs}
\usepackage{multirow}
\usepackage{float}
\usepackage[hidelinks]{hyperref}

\usetikzlibrary{shapes.geometric,arrows,positioning,fit,backgrounds,calc,decorations.pathreplacing,decorations.markings,patterns,circuits.logic.US,matrix,chains}

\definecolor{qubitblue}{RGB}{70,130,180}
\definecolor{controlred}{RGB}{220,20,60}
\definecolor{aigreen}{RGB}{50,150,50}
\definecolor{quantumpurple}{RGB}{128,0,128}
\definecolor{errororange}{RGB}{255,140,0}

\begin{document}

\title{Large Language Models at the Intersection of Software Engineering and Software Security: An Evidence-Centered Structured Survey and Research Agenda}

\author{Wei Lin, Tao Zhou, Zhaofei Xie, and Changgui Hong\\
Nanjing Liancheng Intelligent Technology Group, Nanjing, China\\
\small\texttt{\{linwei, zhoutao, xiezhaofei, hongchanggui\}@chinaliancheng.com}}

\date{}

\maketitle

\begin{abstract}
Large Language Models (LLMs) are moving from code completion toward repository-scale agents that retrieve context, edit files, execute tools, and participate in security-sensitive workflows. The evidence for these systems, however, remains divided between software engineering evaluations centered on functional task completion and software security evaluations centered on vulnerability detection, secure generation, or exploit-oriented validation. This evidence-centered structured survey synthesizes representative work available through May 31, 2026 across software engineering tasks, software security tasks, adaptation mechanisms, artifact granularity, and evaluation design. In addition to a task taxonomy, we introduce an assurance framework that separates functional correctness, security, operational reliability, evidence provenance, and agent authority. The review shows that execution feedback and repository access can substantially improve engineering task completion, but do not by themselves establish security; conversely, static-analysis labels or vulnerability-classification scores rarely establish deployable correctness. We identify recurring validity threats---weak test oracles, duplicated and temporally leaked data, changing agent harnesses, proxy-only security checks, and under-reported budgets and human intervention---and derive a minimum reporting protocol for cross-study comparison. The resulting research agenda prioritizes jointly secure-and-functional benchmarks, repository-scale threat models, calibrated human oversight, longitudinal maintainability evidence, and reproducible agent evaluation. The central conclusion is that model capability should be judged as an assurance case supported by task-appropriate evidence, rather than by a single benchmark score.
\end{abstract}


\section{Introduction}
The emergence of Large Language Models has substantially changed software development research and practice. Models such as GPT-4, CodeLlama \cite{roziere2023code}, StarCoder \cite{li2023starcoder}, and DeepSeek-Coder \cite{guo2024deepseek} can understand, generate, and reason about source code. Their applications span code generation, automated program repair (APR), testing, code review, documentation, and vulnerability analysis \cite{hou2023large}. Practitioner evidence also suggests productivity benefits for selected development activities, although reported gains depend strongly on task, developer experience, codebase context, and evaluation design \cite{gurgul2026state}.

This progress has been accompanied by a proliferation of survey papers, summarized in Table~\ref{tab:surveys}. Hou et al. \cite{hou2023large} provide a systematic mapping of LLM4SE, including some security-related tasks within the broader software engineering landscape. Guo et al. \cite{guo2025benchmarks} focus on benchmarks and solutions for LLM-empowered agentic systems, while Khan et al. \cite{khan2026code} examine code-review benchmarks and evaluation practices. On the security axis, Xu et al. \cite{xu2024llm} review LLM applications across cybersecurity, Zhou et al. \cite{zhou2024llm} focus on vulnerability detection and repair, Sheng et al. \cite{sheng2025llms} synthesize vulnerability-detection techniques, and Basic and Giaretta \cite{basic2026vulnerabilities} trace evidence from insecure generation to detection and remediation. These reviews establish substantial task-specific knowledge, but they leave open a cross-domain question: what evidence is sufficient when a development artifact must be functional, secure, reproducible, and produced by an agent with non-trivial authority?

Our survey adopts an explicitly intersectional perspective. First, it treats \emph{software engineering and software security as co-equal analytical pillars} rather than presenting security only as a subtask of software engineering. Second, it covers the emergence of \emph{agentic systems}, including SWE-bench-oriented architectures \cite{jimenez2024sw}, conversational development paradigms \cite{ge2025vibe}, and autonomous security workflows \cite{xu2024llm}. Third, it connects the two domains by examining how code-generation reliability affects security, how security analysis enters development workflows, and which evaluation criteria are required when functionality and security must be satisfied simultaneously. We do not claim exhaustive coverage of either field; instead, the goal is to provide a structured bridge between them.

\begin{table}[H]
\centering
\caption{Comparison with representative surveys. $\circ$ = covered as a sub-topic; $\bullet$ = primary focus; -- = not emphasized.}
\label{tab:surveys}
\setlength{\tabcolsep}{2.5pt}
\begin{tabular}{lcccccc}
\toprule
\textbf{Survey} & \textbf{Year} & \textbf{SE} & \textbf{Security} & \textbf{Agentic} & \textbf{Benchmarks} & \textbf{Intersection} \\
\midrule
Hou et al. \cite{hou2023large}   & 2024 & $\bullet$ & $\circ$ & --          & $\circ$   & $\circ$ \\
Xu et al. \cite{xu2024llm}       & 2025 & --        & $\bullet$ & $\circ$ & $\circ$   & $\circ$ \\
Zhou et al. \cite{zhou2024llm}   & 2024 & --        & $\bullet$ & --          & $\circ$   & $\circ$ \\
Guo et al. \cite{guo2025benchmarks} & 2025 & $\bullet$ & -- & $\bullet$ & $\bullet$ & -- \\
Gao et al. \cite{gao2025hallucination} & 2025 & $\bullet$ & $\circ$ & -- & -- & $\circ$ \\
Basic and Giaretta \cite{basic2026vulnerabilities} & 2026 & $\circ$ & $\bullet$ & $\circ$ & $\circ$ & $\circ$ \\
\textbf{This survey}             & 2026 & $\bullet$ & $\bullet$ & $\bullet$ & $\bullet$ & $\bullet$ \\
\bottomrule
\end{tabular}
\end{table}

This survey makes four contributions:
\begin{enumerate}[leftmargin=*,label=(\roman*)]
    \item a unified taxonomy that relates engineering and security tasks to artifact granularity, context requirements, adaptation mechanisms, and agent authority;
    \item an assurance-centered framework that distinguishes functional, security, operational, reproducibility, and governance evidence instead of treating task success as a scalar score;
    \item a structured comparison of benchmarks and metrics, including a claim--evidence ladder and a minimum reporting protocol for model and agent evaluations; and
    \item a testable research agenda connecting secure generation, repository-scale analysis, human oversight, maintainability, and standardized cross-task evaluation.
\end{enumerate}

\section{Review Methodology}\label{sec:method}

\subsection{Scope and Research Questions}
This article is an evidence-centered structured narrative and tertiary survey rather than a completed protocol-driven systematic literature review. The search cutoff is May 31, 2026. We prioritize foundational model and benchmark papers, recent systematic reviews, empirical evaluations, and work that directly connects development with software security. The unit of synthesis is not the paper alone, but the claim supported by a particular artifact, dataset, execution environment, metric, and threat model. Because the underlying studies use different model versions, prompts, tools, retry budgets, repository snapshots, and benchmark variants, reported scores are treated as study-specific observations rather than as inputs to a pooled meta-analysis. This scope is deliberately transparent: the review provides a traceable conceptual synthesis and an extensible evidence package, but does not claim exhaustive recall of the literature.

The review is guided by four research questions:
\begin{enumerate}[label=\textbf{RQ\arabic*:},leftmargin=*]
    \item Which software engineering and software security tasks are addressed by LLM-based systems, and what artifacts and context do they require?
    \item How do prompting, fine-tuning, retrieval, and agentic tool use change the capabilities and failure modes observed across these tasks?
    \item Which benchmarks and metrics provide credible evidence, and which validity threats prevent direct cross-study comparison?
    \item Where do engineering and security requirements intersect, and what evaluation and governance mechanisms are needed for integrated agents?
\end{enumerate}

\subsection{Search, Eligibility, and Coding}
Sources were identified through backward chaining from recent domain surveys and targeted searches for task names, adaptation strategies, models, and benchmark names in arXiv, ACM, and IEEE indexes. The protocol defines four concept blocks: language or code models; software engineering; software security; and adaptation or agent mechanisms. A source was retained when it introduced a consequential model, dataset, benchmark, or method; reported empirical evidence relevant to one of the research questions; synthesized a clearly bounded subfield; or exposed a validity threat with direct implications for evaluation. Opinion-only articles, duplicate versions, sources without enough methodological detail to calibrate their claims, and general cybersecurity work without a software artifact were excluded from the core synthesis.

The manuscript's seed corpus is generated automatically from citation keys in the source file and deduplicated by DOI, arXiv identifier, and normalized title. Coding fields include publication status, task family, artifact granularity, programming language, model or system, adaptation strategy, context source, tools, benchmark, split, metric, principal result, limitations, and the SE--security intersection. A separate quality rubric scores research-question clarity, dataset provenance, split validity, configuration reporting, baseline quality, metric validity, uncertainty analysis, leakage analysis, reproducibility, external validity, and security validation. These artifacts support audit and extension; reviewer judgments and screening agreement are intentionally left for real human coding rather than generated automatically.

\subsection{Synthesis and Reproducibility}
The analysis combines task-centered narrative synthesis with structured comparison. Findings are not pooled statistically because the studies differ in models, data, prompts, tool access, and evaluation protocols. Instead, evidence is compared at the level of claims that a benchmark can support: function-level correctness, repository-level task completion, vulnerability classification, secure generation, or operational agent behavior. Positive results are interpreted together with negative findings, proxy limitations, and stated threats to validity. The source package includes a prospective review protocol, machine-readable seed corpus, search-log and extraction templates, a quality rubric, and scripts for merging, deduplication, and descriptive statistics. This package separates what has already been synthesized from what a future exhaustive update must still screen.

\subsection{Claim Calibration}
We calibrate conclusions using four evidence levels. \emph{E1 (illustrative)} consists of qualitative examples, demonstrations, or small curated tasks and supports feasibility claims only. \emph{E2 (controlled benchmark)} adds repeatable datasets, baselines, and task-aligned metrics, supporting comparative claims within a fixed configuration. \emph{E3 (repository and execution)} requires realistic project context, executable tests or exploits, environment capture, and regression checks, supporting system-level claims on the sampled projects. \emph{E4 (operational)} additionally requires longitudinal use, human-effort or cost measurement, incident or maintenance outcomes, and governance controls. Evidence at a higher level does not automatically imply better model quality, but it supports a broader deployment claim. Throughout the survey, we therefore distinguish ``performs on a benchmark'' from ``is reliable or secure in practice.''

The remainder of this paper is organized as follows. Section~\ref{sec:prelim} provides background on LLM architectures and adaptation strategies for code. Section~\ref{sec:se} surveys LLM applications across core software engineering tasks. Section~\ref{sec:security} examines LLM-driven approaches for software security. Section~\ref{sec:benchmarks} compares representative evaluation benchmarks. Section~\ref{sec:challenges} discusses persistent challenges and open problems, and Section~\ref{sec:conclusion} concludes with key findings and future directions.

\section{Background and Preliminaries}\label{sec:prelim}

\subsection{LLM Architectures for Code}
The application of LLMs to code spans three dominant architectural paradigms. \emph{Encoder-only} models, exemplified by CodeBERT \cite{feng2020codebert}, employ bidirectional attention to produce contextual representations suitable for understanding tasks such as vulnerability detection, clone detection, and code search. \emph{Encoder-decoder} architectures, such as CodeT5 \cite{wang2021codet5}, combine bidirectional encoding with autoregressive decoding for tasks including summarization, translation, and repair. \emph{Decoder-only} models support open-ended generation and tool-mediated interaction. Codex provided an early demonstration \cite{chen2021eval}, while CodeLlama \cite{roziere2023code}, StarCoder \cite{li2023starcoder}, and DeepSeek-Coder \cite{guo2024deepseek} established widely used open model families. Architectural labels alone do not determine performance: training data, context length, adaptation, and execution feedback are often equally consequential.

\subsection{LLM Adaptation Strategies for Code}
Four principal adaptation strategies have emerged. \emph{Zero-shot and few-shot prompting} leverages the model's pre-trained knowledge without parameter updates, using natural language instructions or input-output examples to guide code generation \cite{chen2021eval}. \emph{Fine-tuning} adapts models to specific code domains through full parameter updates or parameter-efficient methods such as LoRA and QLoRA, which are particularly important for domain-specific languages \cite{yang2025verilog} and security-sensitive tasks \cite{sheng2025llms}. \emph{Retrieval-Augmented Generation} (RAG) enhances generation quality by retrieving relevant code snippets, documentation, or issue reports from repositories, significantly improving repository-level code generation \cite{tao2025rag}. \emph{Agent-based approaches} decompose complex software tasks into subtasks handled by specialized LLM agents with tool-use capabilities, enabling end-to-end automation of the software development lifecycle \cite{bhati2026agentic}.

\subsection{Task Taxonomy}
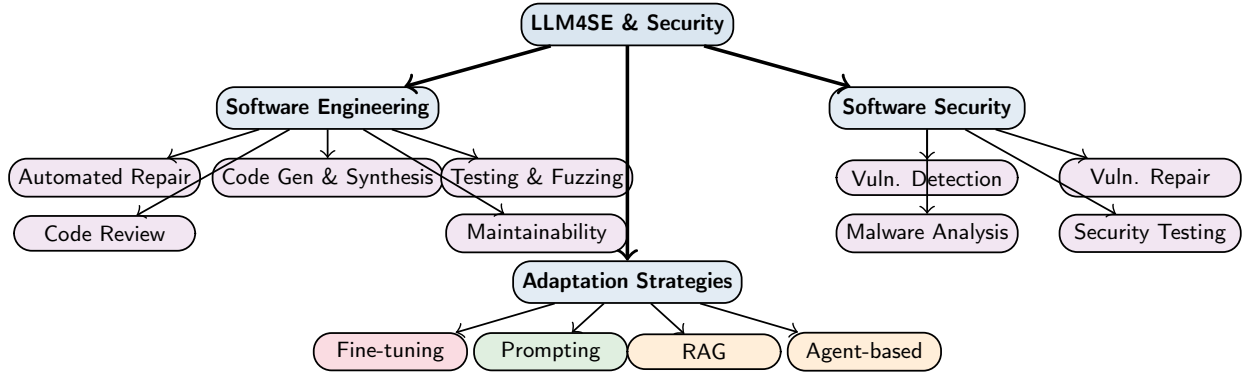
\begin{figure}[H]
    \centering
    \resizebox{\columnwidth}{!}{%
    \begin{tikzpicture}[
        scale=0.55, transform shape,
        level 1/.style={sibling distance=9cm, level distance=1.2cm},
        level 2/.style={sibling distance=2.4cm, level distance=1.0cm},
        l1/.style={rectangle, rounded corners, draw, fill=qubitblue!15, text=black,
            minimum width=2.8cm, minimum height=0.6cm, align=center, font=\sffamily\footnotesize\bfseries},
        l2/.style={rectangle, rounded corners, draw, fill=quantumpurple!10, text=black,
            minimum width=2.6cm, minimum height=0.5cm, align=center, font=\sffamily\footnotesize},
        l3a/.style={rectangle, rounded corners, draw, fill=aigreen!15, text=black,
            minimum width=2.2cm, minimum height=0.5cm, align=center, font=\sffamily\footnotesize},
        l3b/.style={rectangle, rounded corners, draw, fill=controlred!15, text=black,
            minimum width=2.2cm, minimum height=0.5cm, align=center, font=\sffamily\footnotesize},
        l3c/.style={rectangle, rounded corners, draw, fill=errororange!15, text=black,
            minimum width=2.2cm, minimum height=0.5cm, align=center, font=\sffamily\footnotesize},
    ]
    \node[l1,fill=qubitblue!20] (root) at (0,0) {LLM4SE \& Security};
    
    \node[l1] (se) at (-4.3,-1.2) {Software Engineering};
    \node[l1] (sec) at (4.3,-1.2) {Software Security};
    \node[l1] (adapt) at (0,-3.7) {Adaptation Strategies};
    
    \node[l2] (se1) at (-4.3,-2.2) {Code Gen \& Synthesis};
    \node[l2] (se2) at (-7.5,-2.2) {Automated Repair};
    \node[l2] (se3) at (-1.3,-2.2) {Testing \& Fuzzing};
    \node[l2] (se4) at (-7.5,-3.0) {Code Review};
    \node[l2] (se5) at (-1.3,-3.0) {Maintainability};
    
    \node[l2] (sec1) at (4.3,-2.2) {Vuln. Detection};
    \node[l2] (sec2) at (7.5,-2.2) {Vuln. Repair};
    \node[l2] (sec3) at (4.3,-3.0) {Malware Analysis};
    \node[l2] (sec4) at (7.5,-3.0) {Security Testing};
    
    \node[l3b] (a2) at (-3.4,-4.7) {Fine-tuning};
    \node[l3a] (a1) at (-1.1,-4.7) {Prompting};
    \node[l3c] (a3) at (1.1,-4.7) {RAG};
    \node[l3c] (a4) at (3.4,-4.7) {Agent-based};
    
    \draw[->,thick] (root) -- (se);
    \draw[->,thick] (root) -- (sec);
    \draw[->,thick] (root) -- (adapt);
    \draw[->] (se) -- (se1);
    \draw[->] (se) -- (se2);
    \draw[->] (se) -- (se3);
    \draw[->] (se) -- (se4);
    \draw[->] (se) -- (se5);
    \draw[->] (sec) -- (sec1);
    \draw[->] (sec) -- (sec2);
    \draw[->] (sec) -- (sec3);
    \draw[->] (sec) -- (sec4);
    \draw[->] (adapt) -- (a1);
    \draw[->] (adapt) -- (a2);
    \draw[->] (adapt) -- (a3);
    \draw[->] (adapt) -- (a4);
    \end{tikzpicture}
    }
    \caption{Unified taxonomy of LLM4SE and LLM4Security. The hierarchy separates engineering tasks, security tasks, and adaptation strategies; colors distinguish node roles and adaptation families.}
    \label{fig:taxonomy}
\end{figure}

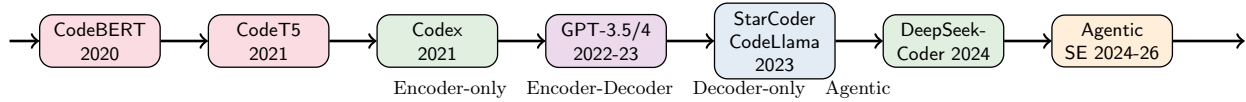
\begin{figure}[H]
    \centering
    \resizebox{\textwidth}{!}{%
    \begin{tikzpicture}[
        scale=0.65, transform shape,
        box/.style={rectangle, rounded corners, draw, fill=qubitblue!15, text=black,
            minimum width=2.0cm, minimum height=0.7cm, align=center, font=\sffamily\footnotesize}
    ]
        \node[box,fill=controlred!15] (cb) at (0,0) {CodeBERT\\2020};
        \node[box,fill=controlred!15] (ct5) at (2.8,0) {CodeT5\\2021};
        \node[box,fill=aigreen!15] (codex) at (5.6,0) {Codex\\2021};
        \node[box,fill=quantumpurple!15] (gpt) at (8.4,0) {GPT-3.5/4\\2022-23};
        \node[box] (sc) at (11.2,0) {StarCoder\\CodeLlama\\2023};
        \node[box,fill=aigreen!15] (ds) at (14,0) {DeepSeek-\\Coder 2024};
        \node[box,fill=errororange!15] (ag) at (16.8,0) {Agentic\\SE 2024-26};
        \draw[->,thick] (-1.5,0) -- (cb);
        \draw[->,thick] (cb) -- (ct5);
        \draw[->,thick] (ct5) -- (codex);
        \draw[->,thick] (codex) -- (gpt);
        \draw[->,thick] (gpt) -- (sc);
        \draw[->,thick] (sc) -- (ds);
        \draw[->,thick] (ds) -- (ag);
        \draw[->,thick] (ag) -- (19,0);
        \node[font=\footnotesize] at (9,-0.8) {Encoder-only\quad Encoder-Decoder\quad Decoder-only\quad Agentic};
    \end{tikzpicture}
    }
    \caption{Timeline of key LLM developments for code (2020--2026). Colors denote architectural paradigms.}
    \label{fig:timeline}
\end{figure}

We organize the LLM4SE+Security landscape along three intersecting dimensions: \textbf{software engineering tasks} -- code generation/synthesis, automated program repair, software testing, code review, and documentation/maintainability; \textbf{software security tasks} -- vulnerability detection, vulnerability repair, malware analysis, and security testing/fuzzing; and \textbf{adaptation strategies} -- the four approaches described above. Figure~\ref{fig:taxonomy} illustrates this taxonomy as a hierarchical tree spanning all three dimensions. Figure~\ref{fig:timeline} situates key developments along a historical timeline from CodeBERT (2020) through the emergence of agentic SE systems (2024--2026).

\section{An Assurance-Centered Framework}\label{sec:assurance}

Existing evaluations often compress a system into one number even though an LLM-based development workflow is a compound intervention. A model receives a context selected by a retriever or agent, acts through a harness, observes tool feedback, and is judged by tests or analyzers executed in a particular environment. We represent an evaluation configuration as
\begin{equation}
\mathcal{E}=\langle A,C,M,T,V,B,H\rangle,
\end{equation}
where $A$ is the software artifact and task, $C$ the available context, $M$ the model and adaptation, $T$ the tools and permissions, $V$ the validators, $B$ the inference and retry budget, and $H$ the human intervention. Two scores are directly comparable only when the differences among these components are either controlled or explicitly analyzed. This representation explains why a model-only label is insufficient for an agent result and why benchmark leaderboards can change when the harness or environment changes.

\subsection{Joint Functional and Security Success}
For an output $y$ on task $i$, let $F_i(y)$ denote satisfaction of functional tests and $S_i(y)$ denote satisfaction of threat-model-aligned security checks. The primary intersectional outcome is
\begin{equation}
J@k=\frac{1}{N}\sum_{i=1}^{N}\mathbb{I}\!\left[\exists y\in Y_i^{(k)}:F_i(y)\land S_i(y)\right],
\label{eq:jointsuccess}
\end{equation}
where $Y_i^{(k)}$ is the set of up to $k$ generated candidates. Reporting $F$ and $S$ separately remains necessary because the same joint score can arise from different failure profiles. The conjunction is nevertheless important: averaging functional and security scores can conceal a catastrophic failure in either dimension. For repair and agent tasks, the conjunction should be extended with regression freedom, environment reproducibility, and policy compliance.

This formulation also separates security from mere absence of analyzer warnings. A security predicate may be supported by dynamic proof-of-vulnerability tests, exploit reproduction, expert review, property checking, or a documented combination of analyzers. The validator must match the claimed threat model. Recent repository-level benchmarks move in this direction by pairing developer tests with dynamic security checks or by evaluating proof-of-concept generation and patching in executable environments \cite{shen2026secrepobench,lee2025secbench,wang2026realsecbench}.

\subsection{Evidence Ladder}
Table~\ref{tab:evidence-ladder} connects evidence design to the strongest defensible claim. The levels are cumulative in realism but not interchangeable: a large controlled benchmark may provide more precise comparative evidence than a small operational deployment, while the deployment provides stronger external-validity evidence. A mature assurance case should therefore combine complementary levels rather than replace one score with another.

\begin{table}[H]
\centering
\caption{Claim--evidence ladder for LLM-based software engineering and security systems.}
\label{tab:evidence-ladder}
\small
\begin{adjustbox}{width=\textwidth}
\begin{tabular}{p{1.0cm}p{3.0cm}p{4.4cm}p{4.0cm}p{3.1cm}}
\toprule
\textbf{Level} & \textbf{Evidence setting} & \textbf{Minimum evidence} & \textbf{Defensible claim} & \textbf{Typical failure} \\
\midrule
E1 & Demonstration or curated examples & Prompt, outputs, manual inspection, and explicit examples & Feasibility on illustrated cases & Cherry-picking and no denominator \\
E2 & Controlled task benchmark & Frozen data and split, baselines, repeated runs, task-aligned metric, configuration disclosure & Comparative performance under the stated benchmark & Leakage, weak tests, and metric mismatch \\
E3 & Repository and executable environment & Versioned repository, build and tests, dynamic or threat-model-aligned checks, regression evidence, tool and budget logs & System effectiveness on sampled projects and environments & Hidden harness effects and irreproducible infrastructure \\
E4 & Operational or longitudinal setting & Real users and workflow, time and cost, maintenance or incident outcomes, access controls, audit trail, and failure escalation & Utility, risk, and sustainability in the observed deployment context & Selection bias, organizational confounding, and rare-event uncertainty \\
\bottomrule
\end{tabular}
\end{adjustbox}
\end{table}

\subsection{Authority and Governance}
Agent authority is a separate evaluation dimension. We distinguish four practical tiers: $G_0$, read-only explanation or retrieval; $G_1$, patch proposal without execution; $G_2$, sandboxed editing and tool execution; and $G_3$, actions that can affect shared repositories, dependencies, credentials, or deployment. Evidence sufficient for $G_0$ is not sufficient for $G_3$. As authority increases, evaluation should add least-privilege tool scopes, network and secret isolation, action logging, approval gates, rollback, and incident handling. This tiering avoids treating all ``agents'' as equivalent and makes governance requirements proportional to potential impact.

\section{LLMs for Software Engineering}\label{sec:se}

\subsection{Code Generation and Synthesis}
Code generation represents one of the most extensively studied LLM4SE applications. Early benchmarks such as HumanEval \cite{chen2021eval} and MBPP \cite{austin2021mbpp} evaluate function-level generation in Python. Reported performance has improved substantially, but HumanEval+ \cite{liu2023human}, which augments the original tasks with many additional tests, demonstrates that solutions passing limited test suites may still fail on edge cases. This distinction between benchmark success and robust functional correctness is central to evaluating generated code.

Repository-level code generation introduces significantly greater complexity. SWE-bench \cite{jimenez2024sw} tasks systems with resolving real GitHub issues spanning entire codebases. The original study reported a 1.96\% resolution rate for its strongest baseline, whereas a recent secondary review reports a result of 78.4\% for a later agentic configuration \cite{bhati2026agentic}. These endpoints illustrate rapid progress, but they should not be interpreted as a controlled longitudinal experiment: benchmark variants, model versions, scaffolding, tool access, and verification procedures changed over time. Table~\ref{tab:swebench} therefore presents the results as illustrative source-reported observations rather than a unified leaderboard.

\begin{table}[H]
\centering
\caption{Illustrative SWE-bench results reported by primary and secondary sources. The rows are not directly comparable because evaluation configurations differ.}
\label{tab:swebench}
\resizebox{\columnwidth}{!}{%
\begin{tabular}{lccp{3.0cm}}
\toprule
\textbf{Source/System} & \textbf{Setting} & \textbf{Resolved (\%)} & \textbf{Interpretation} \\
\midrule
Jimenez et al. \cite{jimenez2024sw} & Original benchmark baseline & 1.96 & Limited repository-level issue resolution \\
Bhati et al. \cite{bhati2026agentic} & Later agentic result reported in review & 78.4 & Stronger models, retrieval, tools, and iterative execution \\
\bottomrule
\end{tabular}
}
\end{table}

Retrieval-Augmented Generation (RAG) is a common mechanism for moving from isolated function generation toward repository-aware synthesis. Tao et al. \cite{tao2025rag} categorize RAG-for-code approaches by generation strategy, retrieval modality, and integration stage. Multilingual code generation remains uneven: Python dominates evaluation, while less represented languages and language-specific toolchains receive substantially less evidence \cite{jiang2026multilingual}. Domain-specific generation, including hardware description languages such as Verilog, further illustrates that syntax, verification infrastructure, and domain constraints materially shape evaluation \cite{yang2025verilog}.

The emergence of ``vibe coding'' -- a paradigm where developers provide high-level intent and iteratively refine LLM outputs through natural language feedback -- represents a shift toward speculative, conversational code generation. Ge et al. \cite{ge2025vibe} synthesize this emerging literature and propose a taxonomy spanning unconstrained iteration to context-enhanced development workflows.

\subsection{Automated Program Repair}
LLM-based APR has become a prominent research direction alongside template-based, search-based, and constraint-based repair. Unlike methods restricted to predefined transformations, generative models can propose diverse patches, but this flexibility also increases the space of plausible yet incorrect repairs \cite{anand2024ai}. A typical pipeline combines fault localization, context construction, patch generation, and validation through compilation, tests, or additional program analyses.

Recent work has explored reinforcement learning and iterative repair, where rewards or feedback encode compilation, test outcomes, and patch properties \cite{masud2026reward}. Such feedback can improve a particular pipeline, but gains depend on test quality and retry budgets. Key benchmarks include Defects4J \cite{just2014defects4j}, BugsInPy \cite{widyasari2020bugsinpy}, and QuixBugs \cite{lin2017quixbugs}. Persistent risks include regression introduction, test-suite overfitting, and semantically vacuous patches that suppress symptoms without satisfying the intended specification.

\subsection{Software Testing and Fuzzing}
LLMs have been applied to test generation and to the automation of testing activities. Execution feedback can help models repair imports, assertions, and setup code, but generated tests may reproduce implementation behavior rather than validate the intended specification. TestGenEval evaluates real-world test authoring, completion, and coverage improvement across Python repositories \cite{jain2024testgeneval}; project-level benchmarks further expose compilation and cascading-context failures that function-level tasks do not capture.

Metamorphic testing, a technique for detecting software faults by verifying relations across program executions, has entered a bidirectional relationship with LLMs. Zheng et al. \cite{zheng2026bidirectional} distinguish metamorphic testing applied to LLMs themselves from LLMs applied to generating metamorphic relations and test oracles.

Fuzzing has been extended by LLM-guided input generation. Rather than relying solely on mutation-based strategies, LLM-guided fuzzers can generate semantically meaningful inputs for structured formats and protocols. Yu et al. \cite{yu2026hitl} review human-in-the-loop fuzzing and identify interaction modes including monitoring, strategy steering, and collaborative exploration. Wienczkowski et al. \cite{wienczkowski2026adaptive} describe a ``structural-adaptive fragmentation'' in which program analysis, DevSecOps, and LLM-based testing remain insufficiently integrated.

\subsection{Code Review and Documentation}
Code review automation has expanded from static analysis and pattern matching to generative review. Khan et al. \cite{khan2026code} document systems that produce comments, estimate change quality, or propose refinements. CodeReviewer provides data and tasks for change-quality estimation, review-comment generation, and code refinement \cite{li2022codereviewer}. CRUXEval instead probes input and output prediction for short programs \cite{gu2024cruxeval}; it is useful for execution reasoning but should not be treated as a direct code-review benchmark.

Code summarization and documentation generation are common LLM applications. Models can produce function-level docstrings, file-level overviews, and architectural descriptions, but linguistic fluency is not evidence of factual consistency. Evaluation should therefore check claims against signatures, call sites, tests, and repository evolution, particularly when documentation describes API behavior or security assumptions.

\subsection{Software Design and Maintainability}
Beyond the coding phase, LLMs are increasingly applied to requirements engineering, architecture design, and design-pattern recommendation. Yu et al. \cite{yu2025aligning} identify a mismatch between academic emphasis on code-level tasks and industrial demand for requirements, architecture, and reliability support. Matias et al. \cite{matias2026llm} similarly identify potential maintainability benefits alongside risks from hallucinations, inconsistent context, and weak lifecycle evidence.

Maintainability claims require different evidence from one-shot correctness. A patch that passes tests today may increase coupling, duplicate logic, weaken an abstraction boundary, or create documentation debt. Useful longitudinal outcomes include subsequent defect density, review and rework effort, code churn, dependency risk, architectural conformance, and the rate at which generated changes are reverted. Current evidence is much stronger for immediate task completion than for these lifecycle outcomes, so productivity and maintainability should not be used as synonyms.

\subsection{Repository-Level Agents and Human Work}
Repository-level agents combine issue interpretation, search, planning, editing, execution, and revision. Their apparent capability is therefore a property of the whole configuration $\mathcal{E}$, not only the base model. Retrieval quality determines which code and documentation become visible; the harness determines editing and tool behavior; the environment determines which tests can run; and retry limits determine how much search is permitted. Evaluations should include component ablations and report failed trajectories, not only the final successful patch.

Human involvement also changes the task. An agent that independently resolves an issue is not directly comparable to a copilot whose user selects files, diagnoses failures, or approves intermediate plans. Human effort should be measured through intervention count, active time, review burden, correction type, and acceptance or rollback decisions. This makes it possible to distinguish automation from productivity support and to identify cases where faster code production merely transfers effort to review, security triage, or maintenance.

\section{LLMs for Software Security}\label{sec:security}

\subsection{Vulnerability Detection}
Detecting security vulnerabilities in source code is one of the most critical and challenging LLM4Security applications. Unlike general bug detection, vulnerability detection requires reasoning about subtle semantic properties -- improper input validation, buffer overflows, injection flaws, and access control violations -- that may span multiple functions or files \cite{xu2024llm}.

For analysis, we group vulnerability-detection approaches into three overlapping families rather than strict chronological generations. \emph{Representation-learning} approaches fine-tune encoder models such as CodeBERT \cite{feng2020codebert} for function-level classification on datasets including Devign \cite{zhou2019devign} and Big-Vul \cite{fan2020big}. \emph{Structure-aware} approaches incorporate control-flow, data-flow, or syntax representations. \emph{Generative reasoning} approaches prompt decoder-only models to produce classifications, explanations, or candidate data-flow traces, sometimes with retrieval and tool support. The families differ in interpretability and context capacity, but all remain sensitive to dataset construction and repository context.

Sheng et al. \cite{sheng2025llms} survey LLM-based vulnerability detection techniques, covering model architectures, adaptation strategies, and evaluation protocols. Zhou et al. \cite{zhou2024llm} organize vulnerability detection and repair approaches from direct prompting and fine-tuning to multi-step reasoning.

Detection granularity ranges from function-level through slice-level to repository-level analysis. Cross-language and multimodal approaches integrate code with commits, issue reports, and security advisories. Representative resources include Big-Vul and DiverseVul \cite{fan2020big,ding2023diversevul}, Devign \cite{zhou2019devign}, ReVeal \cite{chakraborty2021reveal}, and PrimeVul \cite{ding2024primevul}. These datasets differ substantially in labeling quality, class balance, deduplication, and temporal splitting; consequently, scores across them should not be treated as interchangeable. Gao et al. \cite{gao2023how} further show that strong benchmark results do not by themselves establish production readiness.

\subsection{Vulnerability Repair and Patch Generation}
Automated vulnerability repair extends APR to the security domain with heightened correctness requirements: a faulty patch may fail to remove the vulnerability or may introduce a new attack surface. Context-aware repair therefore benefits from information about the vulnerability category, affected paths, and existing mitigations \cite{zhou2024llm}. CVEfixes provides a corpus linking disclosed vulnerabilities to real fixes \cite{bhandari2021cvefixes}; it can support repair research but is not itself a complete repair-evaluation protocol. SecurityEval instead evaluates generated code from a software-security perspective and should be classified under secure code generation rather than vulnerability repair \cite{siddiq2022securityeval}.

Integration of vulnerability repair with general APR pipelines represents an emerging trend. Rather than treating security bugs as a separate category, unified repair frameworks aim to localize a weakness, generate a patch, preserve intended behavior, and validate exploit resistance within one pipeline. SEC-bench, for example, evaluates proof-of-concept generation and vulnerability patching in reproducible task environments \cite{lee2025secbench}; repository-level secure-generation benchmarks similarly combine project context with functional and dynamic security validation \cite{shen2026secrepobench,wang2026realsecbench}. These designs better align the evidence with the claim, although their project, language, and vulnerability coverage remains bounded.

\subsection{Malware Analysis and Threat Intelligence}
Beyond source code analysis, LLMs have been applied to binary-level and behavioral security tasks. Malware detection and classification leverage LLMs' pattern recognition capabilities to identify malicious code patterns, while behavioral analysis uses LLMs to reason about execution traces and system call sequences. Xu et al. \cite{xu2024llm} document LLM applications in threat intelligence, incident response, and security log analysis, noting the emergence of autonomous security agents that orchestrate multi-step security workflows.

Hybrid systems combining traditional program analysis with LLM reasoning represent a particularly promising direction. Static and dynamic analysis provide precise structural information about code and execution, while LLMs contribute semantic understanding and natural language reasoning. This synergy is especially powerful for tasks like vulnerability reachability analysis -- determining whether a detected vulnerability is actually exploitable in a given deployment context.

\subsection{Security Testing and Exploit Generation}
LLM-guided fuzzing augments conventional fuzzers with semantically informed seeds, harness components, or protocol-aware inputs. Wienczkowski et al. \cite{wienczkowski2026adaptive} organize adaptive security testing around structural analysis, DevSecOps integration, feedback-driven fuzzing, LLM-based test generation, and hybrid systems. They identify insufficient integration among program analysis, development pipelines, and LLM-based generation as a central obstacle. Claims of improved coverage must therefore separate the contribution of the LLM from the coverage guidance and execution infrastructure supplied by the underlying fuzzer.

Exploit generation -- constructing inputs that demonstrate practical exploitability -- is inherently dual use. LLM-based security agents may assist with triage, incident analysis, and response orchestration, but evidence for fully autonomous deployment remains limited \cite{xu2024llm}. False positives can increase analyst workload, while false negatives can miss critical threats. Human approval, audit logs, scoped tool permissions, and reproducible execution evidence are therefore important design requirements for security-agent evaluations.

\subsection{Cross-Domain Integration}
The intersection of software engineering and security is most visible at four points in the development workflow. First, generated code should be evaluated for both functional correctness and security properties rather than unit-test passage alone. Second, repair systems should verify that a patch removes the targeted weakness without breaking behavior or creating a new vulnerability. Third, repository-level agents should consume feedback from tests, static analyzers, dependency scanners, and policy checks within the same refinement loop. Fourth, agent authority should depend on risk: read-only analysis, patch proposal, test execution, and deployment are materially different capabilities. These integration points motivate evaluations that report secure-and-functional success, evidence provenance, tool permissions, and the amount of human intervention required.

\section{Benchmarks and Evaluation}\label{sec:benchmarks}

\subsection{Code Generation Benchmarks}
The code generation landscape is served by a diverse ecosystem of benchmarks, summarized in Table~\ref{tab:benchmarks}. HumanEval \cite{chen2021eval} (164 hand-written Python problems) and MBPP \cite{austin2021mbpp} ($\sim$1,000 crowd-sourced Python problems) remain common function-level evaluations, typically measured via pass@k. HumanEval+ \cite{liu2023human} adds substantially more tests to expose solutions that pass weak suites. DS-1000 targets realistic data-science problems across major Python libraries \cite{lai2022ds1000}, while BigCodeBench evaluates complex instructions and diverse library calls \cite{zhuo2024bigcodebench}. LiveCodeBench continuously incorporates newly released contest problems and also evaluates self-repair, execution, and output prediction \cite{jain2024livecodebench}. MultiPL-E translates HumanEval and MBPP to 18 additional programming languages \cite{cassano2022multiple}. At repository scale, SWE-bench \cite{jimenez2024sw} evaluates issue resolution on real GitHub projects. These benchmarks measure different capabilities and should be selected according to the artifact and deployment claim under study.

\begin{table}[H]
\centering
\caption{Key benchmarks for LLM4SE and LLM4Security.}
\label{tab:benchmarks}
\resizebox{\columnwidth}{!}{%
\begin{tabular}{lll}
\toprule
\textbf{Category} & \textbf{Benchmark} & \textbf{Primary Task} \\
\midrule
\multirow{5}{*}{\shortstack{Code\\Generation}} & HumanEval/HumanEval+ & Function-level (Python) \\
 & MBPP & Function-level (Python) \\
 & SWE-bench Verified & Repository-level \\
 & DS-1000 & Data science code \\
 & BigCodeBench/LiveCodeBench & Multi-library/Continuous \\
\midrule
\multirow{3}{*}{Repair} & Defects4J & Real-world Java bugs \\
 & BugsInPy & Real-world Python bugs \\
 & QuixBugs & Multilingual bug fixing \\
\midrule
\multirow{4}{*}{\shortstack{Security\\Detection}} & Big-Vul & C/C++ vulnerability data \\
 & Devign/ReVeal & Function-level detection \\
 & DiverseVul & Diverse vulnerability types \\
 & PrimeVul & Deduplicated temporal evaluation \\
\midrule
Security Repair & CVEfixes & Vulnerability--fix corpus \\
Secure Generation & SecurityEval & Prompt-level security \\
Secure Generation & SecRepoBench/RealSec-bench & Repository-level joint evaluation \\
Security Agents & SEC-bench & Executable PoC and patching tasks \\
Security Testing & OSS-Fuzz & Real-world fuzzing targets \\
\midrule
Understanding & CodeXGLUE/CodeSearchNet & Multi-task/Search \\
\bottomrule
\end{tabular}
}
\end{table}

\subsection{Code Repair and Testing Benchmarks}
Automated program repair is commonly evaluated on Defects4J \cite{just2014defects4j}, BugsInPy \cite{widyasari2020bugsinpy}, and QuixBugs \cite{lin2017quixbugs}. These resources differ in language, project realism, fault complexity, and test adequacy, so a repair result on short algorithmic programs does not establish repository-scale effectiveness. TestGenEval evaluates test authoring and completion on real Python repositories \cite{jain2024testgeneval}. For fuzzing research, OSS-Fuzz offers real open-source targets and operational bug histories \cite{ding2021ossfuzz}; nevertheless, coverage or crash counts remain sensitive to time budgets, harness quality, and the underlying fuzzer configuration.

\subsection{Security Benchmarks}
Security evaluation relies on resources with different purposes. Big-Vul \cite{fan2020big}, Devign \cite{zhou2019devign}, ReVeal \cite{chakraborty2021reveal}, PrimeVul \cite{ding2024primevul}, and DiverseVul \cite{ding2023diversevul} support vulnerability-detection research but differ in construction, labeling, balance, and split strategy. PrimeVul emphasizes deduplication, chronological splitting, and realistic class imbalance, illustrating why scores on older datasets may not transfer to deployment settings. CVEfixes links vulnerabilities to fixing commits and code changes \cite{bhandari2021cvefixes}. SecurityEval contains 130 Python prompts covering 75 CWE types and evaluates generated code from a security perspective \cite{siddiq2022securityeval}.

Newer resources increasingly test functionality and security together. SecRepoBench supplies repository context, developer-written tests, and dynamic proof-of-vulnerability inputs for secure code completion \cite{shen2026secrepobench}; RealSec-bench targets secure generation in real Java repositories and explicitly uses a joint secure-and-functional outcome \cite{wang2026realsecbench}; SEC-bench evaluates agent behavior on executable proof-of-concept and patching tasks \cite{lee2025secbench}. These benchmarks narrow the construct-validity gap, but they should still not be collapsed into a single leaderboard because they evaluate different languages, artifacts, authority levels, and threat models.

\subsection{Metric Limitations and Evaluation Gaps}
Despite the breadth of available benchmarks, several evaluation gaps persist. First, \emph{test-suite adequacy}: pass@k estimates the probability that at least one of $k$ generated programs passes the available unit tests \cite{chen2021eval}; it measures functional correctness only with respect to those tests and does not establish full specification compliance or security. Second, \emph{benchmark saturation}: strong HumanEval scores reduce its discriminative power for newer systems. Third, \emph{data leakage}: training-data contamination may allow models to reproduce memorized solutions. Fourth, \emph{security-specific validity}: accuracy, recall, and false-positive rate do not capture triage cost, exploitability, or time to remediation. Static analyzers are valuable validators but imperfect ground truth; human-validated comparison has exposed substantial per-sample disagreement between analyzer reports and actual security labels \cite{firouzi2026persistent}. Fifth, \emph{agentic evaluation}: multi-step systems \cite{guo2025benchmarks} combine retrieval, reasoning, editing, tool use, and execution, making component attribution and reproducibility difficult. Sixth, \emph{selection by repeated sampling}: best-of-$k$ results can improve while cost, latency, and the number of insecure discarded candidates also increase.

\subsection{Minimum Reporting Protocol}
To make results interpretable and reproducible, we propose the reporting protocol in Table~\ref{tab:reporting}. It applies to both model-centric and agentic studies; fields may be marked not applicable, but should not be silently omitted. The protocol operationalizes the evaluation configuration in Section~\ref{sec:assurance} and allows reviewers to identify whether a reported gain comes from the model, context, harness, tools, budget, validator, or human assistance.

\begin{table}[H]
\centering
\caption{Minimum reporting protocol for LLM-based software engineering and security evaluations.}
\label{tab:reporting}
\small
\begin{adjustbox}{width=\textwidth}
\begin{tabular}{p{2.4cm}p{6.0cm}p{7.1cm}}
\toprule
\textbf{Dimension} & \textbf{Required disclosure} & \textbf{Reason for inclusion} \\
\midrule
Task and artifact & Task definition, artifact granularity, language, repository version, and intended deployment claim & Prevents function-level evidence from being generalized to repositories or operations \\
Model and adaptation & Exact model and date/version, decoding parameters, prompts, examples, fine-tuning data, and retrieval configuration & Separates model capability from prompt, data, and context effects \\
Harness and tools & Agent framework and commit, tools, permissions, network access, sandbox, memory, and stopping rules & Makes agent scores attributable and exposes authority-related risk \\
Budget and human input & Samples, retries, tokens, wall time, monetary or compute cost, interventions, and approval decisions & Prevents hidden search and human labor from being counted as autonomous capability \\
Data and contamination & Dataset version, deduplication, temporal and project split, overlap checks, and unavailable training-data caveats & Calibrates memorization and transfer claims \\
Validation & Functional tests, coverage or mutation evidence, security oracle, threat model, regressions, and manual adjudication & Aligns validators with correctness and security claims \\
Uncertainty and failures & Repeated runs, confidence intervals where meaningful, negative results, failure categories, and sensitivity analyses & Avoids conclusions based only on a favorable run or aggregate score \\
Reproducibility & Code, prompts, logs, containers, dependencies, seeds, generated artifacts, and archival identifier & Enables independent verification after services and repositories change \\
\bottomrule
\end{tabular}
\end{adjustbox}
\end{table}

Table~\ref{tab:strategies} summarizes the dominant adaptation strategies and their applicability across the SE and security task landscape, highlighting how different tasks benefit from different LLM utilization paradigms.

\begin{table}[H]
\centering
\caption{LLM adaptation strategies and their applicability across SE and security tasks. $\bullet$ = primary strategy; $\circ$ = commonly used; -- = rarely applied.}
\label{tab:strategies}
\setlength{\tabcolsep}{4pt}
\resizebox{\columnwidth}{!}{%
\begin{tabular}{lcccc}
\toprule
\textbf{Task} & \textbf{Prompting} & \textbf{Fine-tuning} & \textbf{RAG} & \textbf{Agent-based} \\
\midrule
Code Generation (function)    & $\bullet$ & $\circ$ & $\circ$   & --  \\
Code Generation (repo-level)  & $\circ$   & --      & $\bullet$ & $\bullet$ \\
Automated Program Repair      & $\circ$   & $\bullet$ & $\circ$ & $\circ$ \\
Test Generation               & $\bullet$ & $\circ$   & $\circ$   & $\circ$ \\
Code Review                   & $\bullet$ & $\circ$   & $\circ$   & --  \\
Vulnerability Detection       & $\circ$   & $\bullet$ & $\bullet$ & --  \\
Vulnerability Repair          & $\circ$   & $\bullet$ & $\circ$   & $\circ$ \\
Fuzzing                       & $\bullet$ & --      & --        & $\bullet$ \\
\bottomrule
\end{tabular}
}
\end{table}

\subsection{Cross-Task Synthesis}
Table~\ref{tab:synthesis} consolidates the evidence across task families. Three patterns answer the research questions. First, the required context expands from a prompt and function signature to repository state, build systems, issue history, and organizational policy; adaptation correspondingly shifts from prompting and fine-tuning toward retrieval and tool-mediated agents (RQ1--RQ2). Second, evaluation validity depends on the claimed unit of deployment: function tests cannot justify repository-level or security claims, and detector accuracy cannot substitute for exploitability or analyst-cost evidence (RQ3). Third, the SE--security intersection is not a separate terminal task but a set of constraints on generation, repair, testing, and agent authority (RQ4). A high-quality evaluation should therefore report not only whether a task completed, but also which context and tools were available, which functional and security checks passed, what regressions occurred, and how much human intervention was required.

\begin{table}[H]
\centering
\caption{Cross-task synthesis of artifacts, adaptation strategies, evidence, and principal validity threats.}
\label{tab:synthesis}
\small
\setlength{\tabcolsep}{3pt}
\begin{adjustbox}{width=\textwidth}
\begin{tabular}{p{2.3cm}p{3.0cm}p{3.0cm}p{3.6cm}p{4.2cm}}
\toprule
\textbf{Task family} & \textbf{Artifact and context} & \textbf{Common adaptation} & \textbf{Typical evidence} & \textbf{Principal validity threat} \\
\midrule
Function synthesis & Specification, signature, local tests & Prompting; fine-tuning & pass@k and expanded unit tests & Weak tests, contamination, and limited API realism \\
Repository issue resolution & Repository, issue, build and test environment & Retrieval; planning; tool-using agents & Resolved issues and regression tests & Changing benchmark versions, retry budgets, and hidden infrastructure \\
Program repair & Fault location, buggy code, tests, execution feedback & Fine-tuning; iterative prompting; agents & Plausible and correct patches & Test-suite overfitting and behavior regressions \\
Test generation and fuzzing & Code under test, harness, runtime feedback & Prompting; execution-guided agents & Coverage, mutation score, crashes, valid tests & Oracle weakness and confounding from the underlying fuzzer \\
Review and documentation & Diff, repository context, conventions, APIs & Prompting; retrieval; task-specific tuning & Comment relevance and human judgment & Subjective labels and fluent but unsupported explanations \\
Vulnerability detection & Function, slice, data flow, repository context & Fine-tuning; structure-aware models; retrieval & Precision, recall, F1, localization & Duplicates, temporal leakage, unrealistic balance, and label noise \\
Secure generation and repair & Specification, CWE/CVE context, tests, analyzers & Prompting; fine-tuning; constrained or iterative generation & Functional tests plus security checks & Proxy analyzers, incomplete threat models, and new vulnerabilities \\
Security agents & Multi-tool environment, permissions, logs, policies & Retrieval; planning; tool use; human approval & Task completion, evidence trace, analyst effort & Unsafe actions, irreproducible environments, and unbounded authority \\
\bottomrule
\end{tabular}
\end{adjustbox}
\end{table}

\section{Challenges and Open Problems}\label{sec:challenges}

\subsection{Reliability, Hallucination, and Calibration}
Code hallucinations---outputs that are syntactically plausible but semantically incorrect, insecure, or inconsistent with the repository---remain a fundamental reliability challenge. Gao et al. \cite{gao2025hallucination} relate these failures to training data, autoregressive generation, and insufficient context. At repository scale, the failure surface expands to nonexistent symbols, stale APIs, misunderstood build assumptions, incompatible dependency changes, and patches that address the visible symptom while violating an implicit invariant.

Execution feedback reduces some errors but creates a risk of validator overfitting. An agent may learn to satisfy the observed tests without implementing the intended behavior, or remove an analyzer warning without eliminating exploitability. Reliability therefore requires diverse evidence: specification-derived tests, regression suites, mutation analysis, static and dynamic checks, differential behavior, and targeted human review. Uncertainty should also be calibrated at the level at which action is taken. A confidence statement attached to a fluent explanation is less useful than calibrated probabilities for patch acceptance, vulnerability presence, or the need for escalation.

\subsection{Dataset Realism, Leakage, and Benchmark Governance}
Dataset construction can dominate apparent model progress. Duplicate or near-duplicate functions inflate random-split performance; vulnerability-fixing commits contain unrelated edits; negative examples may be sampled from unrealistically clean code; and labels based on a commit message or analyzer can be noisy. PrimeVul demonstrates the importance of deduplication, chronological splitting, and realistic imbalance \cite{ding2024primevul}. Similar discipline is needed for generative tasks because benchmark problems, tests, solutions, and repository histories may all be present in pretraining or retrieval corpora.

A benchmark should consequently be governed as a versioned scientific instrument. Its release should identify provenance, licenses, filtering, label adjudication, known contamination channels, test coverage, supported claims, and change history. For agent benchmarks, the repository image, dependency registry state, container, network policy, and evaluator must also be frozen. Periodic refreshes can reduce memorization, but they should not silently replace an existing test set because longitudinal comparisons then lose meaning. A stable core, a sequestered evaluation set, and a documented refresh track provide a more defensible compromise.

\subsection{Reproducibility, Cost, and Environmental Drift}
Proprietary model updates, nondeterministic sampling, disappearing APIs, mutable repositories, and evolving package registries make exact replication difficult. Agent evaluations add further variation through search indexes, tool versions, timeouts, concurrency, network access, and retry policies. A result should therefore include both a logical specification of the task and an executable environment whenever licensing permits. Containers alone are insufficient if external services, model snapshots, or package artifacts remain mutable.

Cost is also part of effectiveness. A system that obtains a higher best-of-$k$ success rate through many samples or long repair loops may be unsuitable for routine use. Relevant quantities include tokens, model calls, wall time, compute or monetary cost, tool executions, failed candidates, and human review. Reporting a Pareto frontier over quality, security, cost, and latency is more informative than ranking systems by task success alone. For security triage, analyst time spent dismissing false positives may dominate model inference cost.

\subsection{Adversarial Robustness and Agent Governance}
Security systems face distribution shift and active adversaries. Semantics-preserving renaming, control-flow restructuring, dead-code insertion, misleading comments, poisoned retrieval documents, and tool-output manipulation can alter model judgments. Repository-level agents additionally process untrusted files and issue text that may contain prompt injections. Evaluations should distinguish robustness to natural code evolution from robustness to deliberate manipulation and should test both the model and the surrounding tool chain.

Governance becomes essential when agents can execute commands or modify shared assets. The authority tiers in Section~\ref{sec:assurance} imply progressively stronger controls: isolation and content provenance for read-only analysis; mandatory diff review for proposed patches; sandboxing and egress restrictions for tool execution; and explicit approval, rollback, and audit requirements for changes to production or security infrastructure. Safety refusal rates alone do not measure these system properties. The relevant question is whether unsafe actions are prevented, detected, contained, and recoverable under the stated threat model.

\subsection{The Academic--Industry Evidence Gap}
Yu et al. \cite{yu2025aligning} identify mismatches in task emphasis, scale realism, and integration. Academic work is concentrated on code-level tasks with automatically checkable outcomes, whereas industrial adoption must also address requirements, architecture, migration, reliability, compliance, ownership, and coordination. Million-line repositories, private dependencies, legacy build systems, organization-specific conventions, and regulated data cannot be represented fully by small public benchmarks.

The gap is not solved simply by obtaining proprietary data. Industrial studies also require credible counterfactuals: comparison teams or periods, developer-experience stratification, workflow instrumentation, and measurement of downstream review, incidents, and maintenance. Matias et al. \cite{matias2026llm} emphasize that long-term maintainability remains uncertain. Longitudinal studies should follow generated changes beyond merge time and should report rejected and reverted outputs, not only accepted suggestions.

\subsection{Threats to the Validity of This Survey}
\emph{Selection validity} is limited because this is a structured narrative review built from a traceable seed corpus rather than a completed exhaustive search and dual-reviewer screening process. Relevant work may be absent, and highly visible English-language papers and preprints may be over-represented. The protocol and templates make this limitation auditable but do not remove it. \emph{Construct validity} is threatened by inconsistent uses of terms such as correctness, agent, autonomy, vulnerability detection, and secure generation; the taxonomy and assurance framework reduce but cannot eliminate these differences. \emph{Conclusion validity} is limited by heterogeneous tasks and configurations, which is why source-reported scores are not pooled. \emph{External validity} remains constrained by the dominance of a small set of languages, open-source repositories, curated datasets, and short evaluation windows. \emph{Temporal validity} is unusually fragile because models, APIs, scaffolds, and leaderboards change rapidly. The cutoff date and benchmark versions should therefore accompany any reuse of the conclusions.

\subsection{A Testable Research Agenda}
Table~\ref{tab:agenda} translates broad directions into falsifiable questions and minimum evidence. The agenda favors studies that isolate mechanisms, preserve failure evidence, and jointly evaluate engineering value and security risk.

\begin{table}[H]
\centering
\caption{Research agenda derived from the evidence gaps.}
\label{tab:agenda}
\small
\begin{adjustbox}{width=\textwidth}
\begin{tabular}{p{3.0cm}p{5.7cm}p{7.0cm}}
\toprule
\textbf{Direction} & \textbf{Testable research question} & \textbf{Minimum credible evaluation} \\
\midrule
Joint secure generation & Does an intervention improve $J@k$ rather than shifting failures between functionality and security? & Paired functional and threat-model-aligned tests; separate $F$, $S$, and joint outcomes; fixed sampling budget; regression and failure analysis \\
Repository context and agents & Which gains come from retrieval, the model, the harness, tools, or additional retries? & Versioned repositories and environments; controlled component ablations; equalized budgets; trajectory and tool logs; multiple projects \\
Verifier ensembles and human feedback & When do tests, analyzers, dynamic exploits, and expert review provide complementary rather than correlated evidence? & Human-adjudicated subset; precision, recall, calibration, disagreement analysis, analyst time, and sensitivity to the chosen oracle \\
Risk-bounded autonomy & How does increasing authority change utility, unsafe-action probability, and oversight cost? & Evaluation across $G_0$--$G_3$ tiers; adversarial tasks; least-privilege controls; approval and rollback logs; incident-oriented metrics \\
Lifecycle maintainability & Do LLM-authored changes reduce total engineering effort without increasing future defects or debt? & Longitudinal repository study; matched controls; review and rework time; defects, reversions, churn, and architecture measures \\
Continuous and multilingual evaluation & Do methods transfer across time, languages, ecosystems, and newly disclosed weakness patterns? & Temporal and project-held-out splits; multiple language/toolchain families; benchmark refresh protocol; contamination audit and uncertainty reporting \\
\bottomrule
\end{tabular}
\end{adjustbox}
\end{table}

\section{Conclusion}\label{sec:conclusion}
This survey connected software engineering and software security through a shared taxonomy, an explicit evaluation configuration, and an assurance-centered view of evidence. The synthesis answers the four research questions as follows. For RQ1, the tasks range from function-level generation and classification to repository issue resolution, security patching, fuzzing, and tool-mediated analysis; the decisive difference is often the artifact and context available rather than the task label alone. For RQ2, prompting and fine-tuning remain important for bounded tasks, while retrieval, execution feedback, and agents become more consequential as repository context and iteration increase. These mechanisms change both capability and failure modes, and their effects cannot be attributed to the base model without controlled ablations.

For RQ3, credible evidence must match the deployment claim. Unit tests support only test-relative functional claims; vulnerability labels support classification claims; and repository or operational claims require versioned environments, realistic context, regression checks, security oracles, budgets, and human-effort evidence. The proposed configuration $\mathcal{E}$, evidence ladder, and minimum reporting protocol make these requirements explicit. For RQ4, security is not a downstream task appended to development. It is a constraint on generation, repair, testing, review, and agent authority. Joint outcomes such as $J@k$ should therefore accompany separate functional and security results.

The literature supports cautious optimism rather than unconditional automation. Repository access and iterative tool use can improve task completion, yet they also introduce harness dependence, larger attack surfaces, higher cost, and reproducibility challenges. Static analyzers, tests, and LLM judges are useful components of an assurance case but are not individually sufficient ground truth. Long-term maintainability, multilingual transfer, adversarial robustness, and the organizational cost of oversight remain under-evidenced.

The central research priority is consequently not another isolated leaderboard, but a cumulative evidence infrastructure: versioned benchmarks, executable environments, contamination-aware splits, joint functional and security validation, transparent agent permissions, complete trajectory and budget logs, and longitudinal human-centered outcomes. Progress should be claimed in proportion to the evidence. Under this standard, an LLM-based software system is trustworthy not because it produces a plausible patch or achieves a high scalar score, but because its behavior is supported by a reproducible, threat-model-aligned, and governance-aware assurance case.



\bibliographystyle{unsrt}
\bibliography{ref}

\end{document}